\documentclass[11pt]{article}

\usepackage[margin=1in]{geometry}
\usepackage[T1]{fontenc}
\usepackage{lmodern}
\usepackage{microtype}
\usepackage{amsmath}
\usepackage{amssymb}
\usepackage{booktabs}
\usepackage{tabularx}
\usepackage{longtable}
\usepackage{array}
\usepackage{xcolor}
\usepackage{graphicx}
\usepackage{hyperref}
\usepackage{enumitem}
\usepackage{listings}

\hypersetup{
  colorlinks=true,
  linkcolor=blue!55!black,
  citecolor=blue!55!black,
  urlcolor=blue!55!black
}

\setlist[itemize]{leftmargin=1.5em}
\setlist[enumerate]{leftmargin=1.5em}
\newcolumntype{Y}{>{\raggedright\arraybackslash}X}

\newcommand{\DeltaAUC}{\Delta_{\mathrm{AUC}}}
\newcommand{\manualtablecaption}[1]{%
  \refstepcounter{table}\par\smallskip Table~\thetable: #1\par
}
\lstdefinestyle{noteexcerpt}{
  basicstyle=\ttfamily\scriptsize,
  breaklines=true,
  columns=fullflexible,
  keepspaces=true,
  frame=single,
  framerule=0.3pt,
  xleftmargin=0.5em,
  xrightmargin=0.5em,
  aboveskip=0.75em,
  belowskip=0.75em
}

\title{Specification Oracles}
\author{Atticus Cull, Justin McCarthy\\
Diffusion\\
Redwood City, California, USA}
\date{September 3, 2026}

\begin{document}
\maketitle

\begin{abstract}

Specifications face a basic tradeoff: leave details out, and important questions go unanswered; record every detail separately, and the specification becomes large and prolix. We investigate whether a language model can serve as a compact, living specification oracle by learning facts about a target and answering questions about it directly. We compare two ways of storing the learned facts: external text notes and changes to the model’s weights. Across four families of 596-fact worlds and two Qwen2.5 model sizes, weight-only oracles benefited substantially more from structure: with the 7B model, their accuracy integrated across storage capacities was 18.5 percentage points higher on structured than unstructured worlds, compared with 1.1 points for note-sheet oracles. This advantage came at a substantial storage cost, with the smallest adapter requiring approximately 175 KiB compared with a maximum note budget of 16 KiB. Adapted weights therefore exploited latent structure more successfully, while external notes required substantially less object-specific storage.
\end{abstract}

\section{Introduction}

\subsection{Specification Format and Compression}

We are motivated by the problem of specification, the task of extracting all of
the important contours, behaviors, and properties of some target object. Of
course, before writing any sort of specification, one has to decide on
the format. The format dictates both how the specification should be interpreted
and what methods of description the specification has access to. This
choice of format is critical: think of the difference in using vector versus
raster graphics, for instance. This paper will be focused in
particular on the implications of format on the needed size of an accurate
specification for an object. On the one hand, fixing a format and varying the
objects, we can see this as a measurement of size/complexity of the object. On
the other hand, fixing an object and varying the format of specification, we get
a measurement of compression efficacy of a format for a given type of object.
Particularly nice or short specs save space by offloading as much information
content as possible onto the priors and primitives of the format. A PNG of a
circle needs to select every pixel individually on the boundary because the PNG
format does not know what a circle is. However, it is extremely simple for an
SVG to represent a circle because it is built in to the format.

\subsection{Specification Oracles}

In hopes of building an all-purpose format for specification, we look towards
neural networks as general purpose knowledge acquirers. A substantial line of
work has established that the parameters of a language model can act as a store
of factual knowledge that is queryable directly in natural language
\cite{roberts2020knowledge,heinzerling2021kb}, and more recent work has begun to
quantify how many bits of such knowledge a network can hold, on the order of a
few bits per parameter \cite{allenzhu2025physics,morris2025memorize}. Our idea is
to build a model that serves as a living specification for an object.
Whereas with a typical plaintext specification, one answers simple fact
questions they might have about the object by reading the relevant
specification, the oracle can just answer those questions directly. Of course,
to have any hopes of a model being accurate at this oracle task for a specific
object, that model has to learn about the object first. We investigate two methods
of learning for such an oracle, accompanied by measurements of size for the
resulting oracles.

In both cases, we start with some base model \(M\). This takes the place of a
plug-and-play way to control the system of priors of the format of
specification. The first method, which we will call a note-sheet oracle, keeps
the model \(M\) completely the same, but allows it read/write access to a file
of fixed size. During recall, \(M\) is given a fresh context with just this note
sheet. With this method, the size of the note sheet is used as the
reported spec size. Importantly though, the note sheet alone is not the spec.
The model is still used as the decoder along with the note sheet --- otherwise
the note sheet has no meaning. The second method, which we will call a
weight-only oracle, trains \(M\) with inserted LoRA parameters with a given
fixed budget of bits. Recent work studies LoRA adapters as a modular
knowledge-memory medium and measures how many bits they can write into a frozen
base model \cite{back2026lora,tan2026bits,ye2026cram}. Here we are measuring not
just the raw number of trainable parameters, but also the precision of the
parameters.

The last ingredient for the experiment is the objects of study. For
this, we use generated virtual worlds, which are in practice a list of
relationships between rooms, regions, key items, factions, etc. We use four
different methods of world generation that provide different kinds of structure
to the worlds. This is to check that the oracles are doing some sort of
compression when there is more structure within the worlds. Correlated facts
that share latent structure are known to be memorized more readily than
independent ones \cite{lu2024scaling}, and our aim is to test whether that
structure translates into a smaller specification. Unlike this body of work,
which largely measures the capacity of a fixed model or adapter to absorb
arbitrary facts, we hold the base model fixed and vary the structure of the
object being specified, asking whether the required specification shrinks when
the object has exploitable structure rather than simply how much a network can
memorize \cite{morris2025memorize,tan2026bits}.

\section{Measuring Structural Compression}

\subsection{Oracle State and Evaluation}

Before continuing, we'll define the terms at play, and the measurement
being taken in the experiment. We will consider an oracle to be the combination of a frozen base model $M$, together with an internal state $S$ which is object-specific, and procured via a learning harness. In the case of the note sheet approach, the state is a single UTF-8 file available to the base model at query time; in the case of the weight-only approach, the state is a set of trained LoRA parameters. The state takes a different medium in each case, and hence our primary interest will be in comparing the absolute accuracy curves within each medium. We use \(\DeltaAUC\), the difference between \(AUC\) in structured vs independent worlds within a single medium, as the main cross-condition summary.

\subsection{Capacity-Accuracy Curves}

Each configuration of the experiment involves: a world $O$ pseudorandomly generated by a generator and a seed, and an accompanying comprehensive set of questions $Q$ about $O$; a frozen base model $M$; and an internal state $S$ with a fixed budget in terms of either note sheet size or fixed parameter-precision LoRA adaptation layer count. For each configuration, we measure the AUC of the final trained oracle on the question set $Q$. The query decoding is fixed in the evaluation harness, so we report uncertainty across the world seeds instead of repeated samples of the same trained oracle.

For a fixed oracle state medium, base model \(M\), and world generator family $G$, we look at the AUC curve as we vary the internal state capacity. We are interested in the comparison of these curves between world families with structure and the independent world generator to see if the oracle is capable of taking advantage of structure to compress better. For the weight-only medium, we can calculate the total amount of storage the adapted parameters in bytes use as a useful storage accounting. However, the ranges of bytes used for the note-sheet and for the weight-only approaches are entirely separate, so we do not put them on the same axis.

\section{Synthetic Worlds and Oracle Implementations}

\subsection{World Generators}

We'll now dive deeper into the specifics of the objects and the learning
harness. Starting with the objects, each world consists of a list of
rooms, regions, factions, important items, and door relations. The rooms have
some attributes such as lighting, material type, and hazard.
Additionally, there are relationships between these, such as which faction owns
which room, what room each item is in, what key unlocks which door, etc. In
total, regardless of the world generation method, the world has an associated
list of 596 facts about it.

\begin{itemize}
  \item \textbf{Independent:} room attributes and door attributes are drawn from
  the appropriate answer vocabularies with independent pseudorandom choices.
  The region/room/item schema is still shared, but there is no intended latent
  rule tying most answer values together.
  \item \textbf{Clustered:} each region receives a theme, and rooms/doors in
  that region reuse the theme's hazard, material, lighting, symbol, owner,
  key, and color values. The compressible rule is mostly ``remember the region
  theme, then apply it repeatedly.''
  \item \textbf{Factorized:} different fields are controlled by different axes:
  hazards by region theme, materials by architecture, lighting by room type and
  region index, symbols/owners by faction and room type, door keys by door
  position, and door colors by door position plus region index. This is less
  visually obvious than clustering because the model has to compose several
  deterministic factors.
  \item \textbf{Hierarchical:} regions have both a theme slot and a lineage
  profile. Some facts follow the region theme, some follow the lineage profile,
  and some depend on local room or door position. This creates repeated
  structure at multiple levels rather than one flat region cluster.
\end{itemize}

\subsection{Question Splits}

During the learning loop, the oracle is able to see facts about the world, and
is tested with questions about the world. For each fact, there are multiple
different forms of question asked about it. The questions are split into three
pools: train, dev, and test. The training questions are used to give the oracle
feedback during the learning process. The dev questions are used to
assess the oracle during training to determine when to stop training.
The test questions are used after training is complete to evaluate the
oracle. The purpose of the split is not to limit fact exposure; the learning
model should have access to every fact. Rather, the questions are split to not
leak question wording and avoid overtraining.

\subsection{Note-Sheet Oracle}

For the note-sheet oracle, the model receives a compact fact table and
its current fact sheet and iteratively is prompted to generate
revisions to its note sheet; lastly, it is given feedback on
which questions it got correct with its updated note sheet. Importantly, the
note sheet is governed with a hard limit on size. If the oracle attempts to
write over its budget, those changes are truncated to fit the budget. After it is
determined that the oracle has saturated its learning, the note sheet is frozen
as final and evaluated. Note that whenever the oracle is being
evaluated, including during the learning phase, the oracle is given a fresh
context with only the note sheet.

\subsection{Weight-Only Adapter Oracle}

For the weight-only oracle, we train the oracle using rank-1, all-module LoRA
over a prefix of layers. The adapter is trained on the training questions
as well as the object specific fact material. The capacity of the oracle is
controlled primarily by the number of
adapted layers, since rank is extremely coarse, and even rank-1 introduces
plenty of parameters. Otherwise, the harness is similar to the note-sheet
oracle. Once saturation is hit as determined by the dev questions, the new
parameters are frozen and the model is evaluated.

\subsection{Oracle Contract}

The two oracle systems differ in what object-specific state is
stored, how that state is produced, and what information is available at query
time. Table~\ref{tab:oracle-contract} summarizes the contract for each system.
This distinction matters because the experiment compares complete oracle
harnesses, not storage media in isolation.

\begin{table}[ht]

\centering
\small
\begin{tabularx}{\linewidth}{@{}lYY@{}}
\toprule
Property & Note-sheet oracle & Weight-only adapter oracle \\
\midrule
Base decoder & Frozen Qwen model & Frozen Qwen model plus active adapter \\
Stored program & UTF-8 note text & Trainable LoRA parameters \\
Swept capacity & Allocated byte budget & Adapted layer prefix / trainable parameters \\
Reported final storage & UTF-8 bytes in the final note & Trainable parameters \(\times\) bytes per stored parameter \\
Study input & Compact fact table, current note, feedback & Object-specific QA/fact training data \\
Selection signal & Dev accuracy & Dev accuracy \\
Evaluation input & Final note plus one question & One question, no note \\
\bottomrule
\end{tabularx}
\caption{Oracle contract for the two systems.}
\label{tab:oracle-contract}
\end{table}

\section{Experimentation Setup}

\subsection{7B Adapter Sweep}

We move on to the range of hyperparameters across which we conducted the experiment.
All of the world generators use seeded pseudorandomness, making
them reproducible. We started first with Qwen2.5-7B-Instruct as the
base model. With the weight-only oracle, we ran instances with adapted layer
counts 1, 2, 4, 8, 16, and 28 and across three seeds and each world
generation type. The training used a total of 64 dev questions per checkpoint,
and the final accuracy was measured with 139 test questions.

\subsection{7B Note Sweep}

With the note-based oracle, we ran instances across note sheets with
allotted budgets of 1KiB, 2KiB, 4KiB, 8KiB, and 16KiB and
across eight seeds and each world generation type. As this method was
computationally more expensive, we opted for 32 dev questions and 100 test
questions instead.

\subsection{14B Scale-Up Sweeps}

Finally, we ran sweeps on Qwen2.5-14B-Instruct, a model with twice the
parameters as the initial runs. For the note-sheet runs, we used the same set of
budgets and seeds as before, and 32 dev questions and 100 test questions. For
the weight-only runs, we again ran three seeds and all the same layer
depth counts with an additional sweep with a layer depth of 48. Again, due to
tractability concerns, we limited the number of test questions to 32.

\par
The following table summarizes these hyperparameters.

\smallskip
\begin{center}
\small
\begin{tabularx}{\linewidth}{@{}lllYccc@{}}
\toprule
Model & Oracle & Condition & Capacity grid & Seeds & Dev \(n\) & Test \(n\) \\
\midrule
7B & Adapter & Semantic & 1, 2, 4, 8, 16, 28 layers & 3 & 64 & 139 \\
7B & Note & Semantic & 1, 2, 4, 8, 16 KiB; 32 KiB diagnostic for 3 seeds & 8 & 32 & 100 \\
7B & Note & Randomized labels & 1, 2, 4, 8, 16, 32 KiB & 3 & 32 & 100 \\
14B & Adapter & Semantic & 1, 2, 4, 8, 16, 28, 48 layers & 3 & 32 & 32 \\
14B & Note & Semantic & 1, 2, 4, 8, 16 KiB & 8 & 32 & 100 \\
\bottomrule
\end{tabularx}
\end{center}

\subsection{Note Harness Caveats}

We mention some caveats with the note-sheet harness that we ran into from
initial tests. Just raw prompting the learner to produce changes to
the note sheet resulted in empty notes, since at the start of the learning curve
an empty note can sometimes outperform a small note. Our fix was to use chat
templates to ensure that at least something was output. We also encourage the
note writer with dense-encoding instruction. This prevents the models from
wasting budget towards cleanly laid out notes, since space is a top concern.

\subsection{AUC and Structure Contrast}

For a given seed and world family, our analysis integrates accuracy over log capacity to
produce an AUC value. For a given seed, the primary structure contrast is the difference
between the average AUC values across the three structured families, and the AUC value
of the independent world.
\[
\DeltaAUC =
\frac{1}{3}\left(\mathrm{AUC}_\mathrm{clustered} + \mathrm{AUC}_\mathrm{factorized} + \mathrm{AUC}_\mathrm{hierarchical}
\right) - \mathrm{AUC}_{\mathrm{independent}}.
\]
We also compute 95\% t-intervals across seed-level \(\DeltaAUC\) values with each seed being a single sample.

\section{Structural Compression and Storage Results}

\begin{figure}[tbp]
\centering
\includegraphics[width=\linewidth]{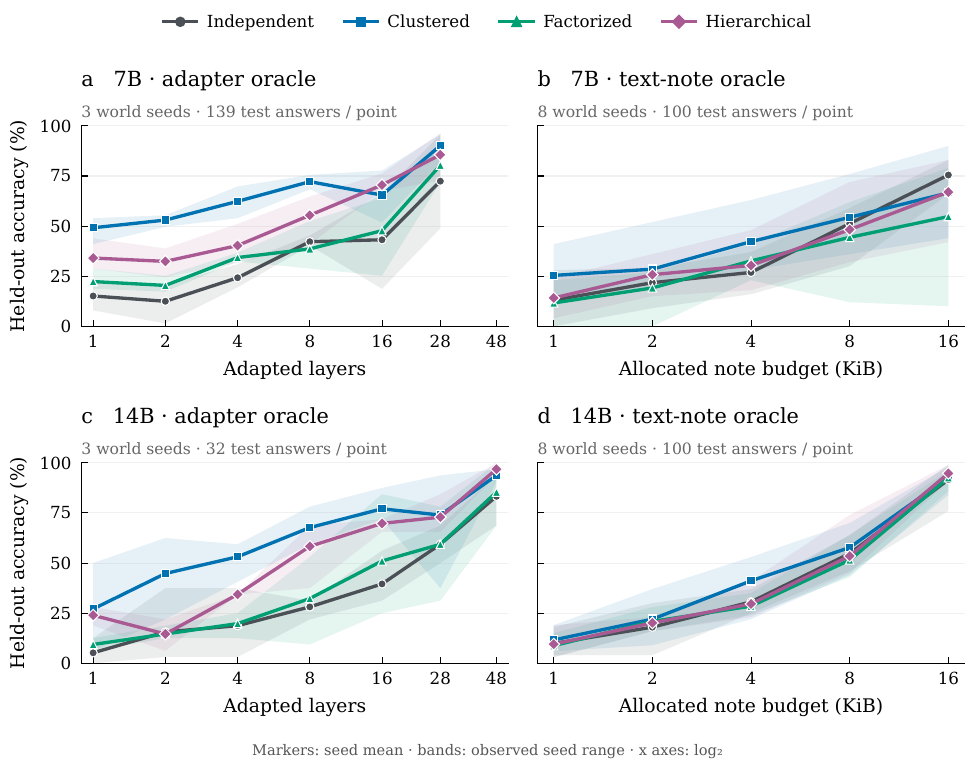}
\caption{Held-out accuracy across capacity for
the four semantic sweeps. Adapter capacity is adapted layer count; note capacity
is allocated UTF-8 note budget. Markers show seed means, and shaded bands show
the observed seed range, not confidence intervals.}
\label{fig:capacity-curves}
\end{figure}

\subsection{Adapter Oracles Exploit Structure}

Our strongest result is the performance of the weight-only oracle. With
$\DeltaAUC = +18.5$~pp, a 95\% t-CI of $[+15.2, +21.8]$,
and all three seeds positive, this is evidence towards the efficacy of
trainable weights as a means of compression of structure in an object. The type
of structure matters though. While $\DeltaAUC$ was large for
Clustered and Hierarchical, it was not as large for Factorized (but
still positive). Looking at the accuracies themselves, the model only needed a
few layers to reach high accuracy for structured world families. The factorized family is informative because its exploitable structure is
not a single region-level lookup. Some fields depend on region theme, others on
architecture, room type, faction, door position, or combinations of those
variables. The smaller performance gap for factorized worlds suggests that the adapter oracle is
better at exploiting repeated clusters than at discovering and applying several
interleaved deterministic rules.

\begin{figure}[tbp]
\centering
\includegraphics[width=\linewidth]{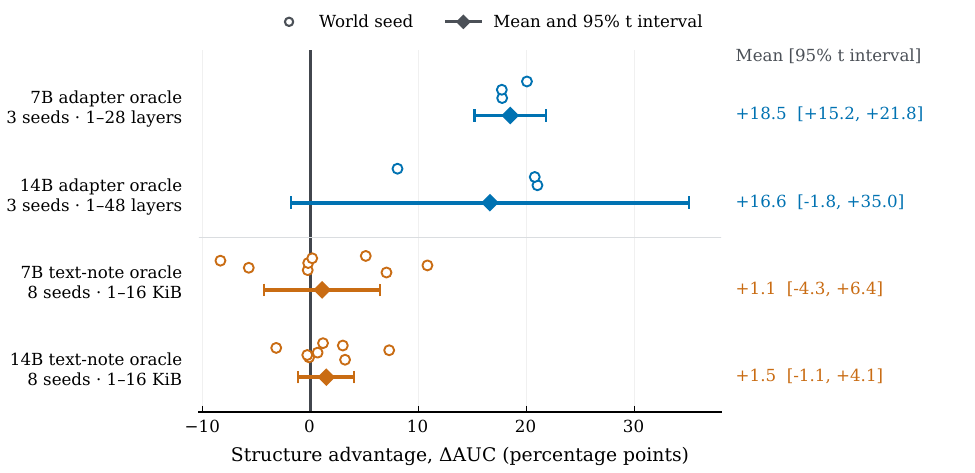}
\caption{Structure advantage,
\(\DeltaAUC\), computed within each model/oracle system as structured-family
AUC minus independent-family AUC. Hollow points are seed-level contrasts; filled
diamonds and bars show the mean and seed-level 95\% Student-\(t\) interval.
This is the cleanest visual summary because it does not require adapters and
notes to share a byte axis.}
\label{fig:structure-advantage}
\end{figure}

\subsection{Note Oracles Show Little Structure Advantage}

On the other hand, the note-sheet approach did not show any significant
advantage on structured worlds. The eight seeds were split four and four on
having positive versus negative values of $\DeltaAUC$, with a
95\% t-CI of $[-4.3, +6.4]$. This does not mean the note-sheet oracle is useless, as accuracy still rises with larger note budgets. Rather, this rise is similar on independent and structured worlds. It is also worth noting that past a certain point,
allocating more budget in the note-sheet did not actually help the oracle, with
notes plateauing at 10-13KB regardless.

\subsection{Storage Ranges Do Not Overlap}

Comparing the two media's space efficiency, we cannot put them directly against each other, as the range of storage swept for each medium do not overlap. The largest note-sheets were allowed a budget of 16KiB,
while even the smallest weight-only oracle, one layer of rank-1 LoRA stored
with 2-byte precision floats, already uses approximately 175\,KiB. What we can take away is that the note-sheet oracle can achieve high raw accuracy with a much smaller allocated byte range, especially in the larger 14B base model case. On the other hand, only the LoRA based oracle approach was able to achieve noticeable boosts in accuracy on structured worlds.

\begin{figure}[tbp]
\centering
\includegraphics[width=\linewidth]{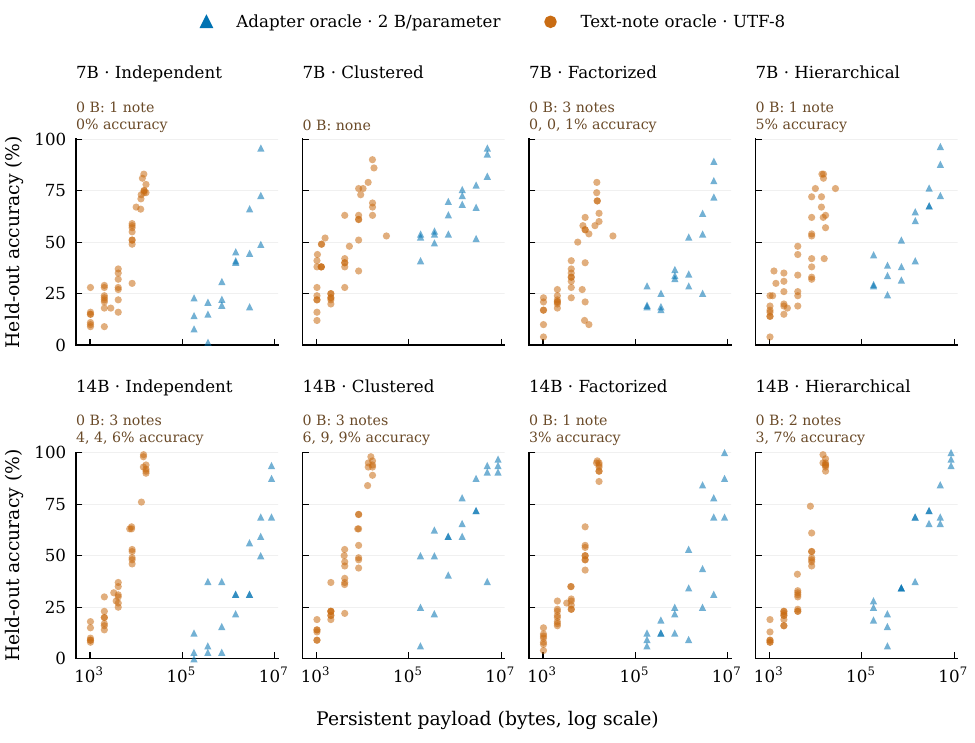}
\caption{Storage accounting. Notes use final UTF-8 note length; adapters use
trainable parameters \(\times\) 2 bytes. The plotted ranges show why a single
accuracy-per-byte ratio would be misleading: note and adapter observations are
separated by roughly an order of magnitude in persistent payload size.}
\label{fig:storage-accuracy}
\end{figure}

\subsection{Qwen2.5-14B Scale-Up}

Now, we look at the results for Qwen2.5-14B as a scale-up check. For the
weight-only oracle, we see a similar $\DeltaAUC$ to the 7B run
of $+16.6$~pp with all seeds positive ($+21.0, +20.8, +8.1$).
This replicates the qualitative split between the structured and independent worlds. However, due to the aforementioned test question limitation, the 95\%
t-CI is much wider: $[-1.8, +35.0]$. Regardless, it is evidence that the
method works with stronger base models, albeit weaker evidence due to the limitations.
The note-based scale-up gives a stronger comparison, with eight seeds and 100 test questions per seed. Once again, we got a low $\DeltaAUC$ of
$+1.5$~pp, with not all seeds having positive values and a 95\% t-CI of
$[-1.1, +4.1]$. Again, this replicates the same behavior as the 7B note-sheet runs.

\subsection{Inspected Notes Look Mostly Memorized}

Taking a look at the actual note sheets that were produced, it is not surprising
that the note-sheet oracle is not performing better on structured worlds.
Across both the structured and unstructured world families, the generated notes
appear to just be a raw list of memorized facts. For example, the 7B clustered run at seed 7 and 16 KiB
begins with repeated entries such as ``\texttt{South Basalt Garden Shrine 1
\(\mid\) hazard=ash bloom}'' and ``\texttt{Broken Basalt Garden Archive 2
\(\mid\) hazard=ash bloom}'', even though the repeated value is region-level
structure in this world. No latent structure is being
abstracted. Hence, world family has no effect on the final size of the note
sheet, just fact count - which in our case is a constant.

\begin{figure}[tbp]
\centering
\includegraphics[width=\linewidth]{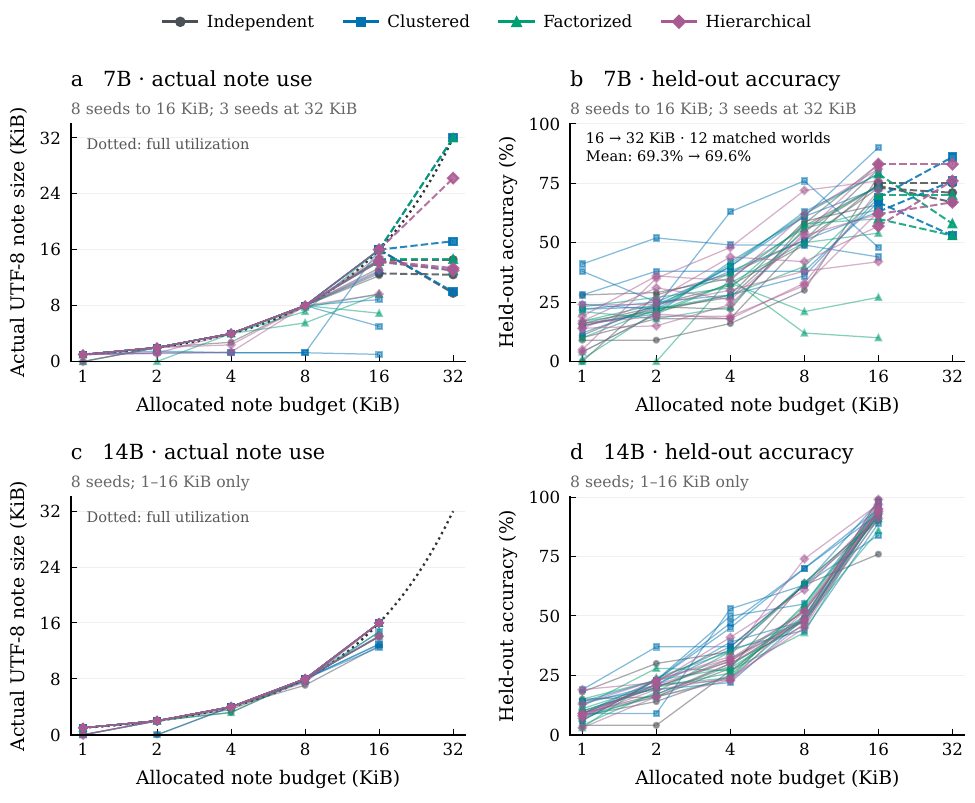}
\caption{Note-utilization diagnostic. The 7B 32 KiB extension
uses only three matched seeds and should be treated as exploratory; among those
12 matched worlds, mean accuracy changes only from 69.3\% to 69.6\%, with
improvements, stalls, and regressions all present. The 14B note sweep uses the
16 KiB budget more fully and reaches high raw accuracy without producing a
large structure advantage.}
\label{fig:note-utilization}
\end{figure}

\section{Interpreting the Compression Tradeoff}

\subsection{Main Interpretation}

Our primary takeaway is that generated structure can effectively be learned by
an oracle model, as demonstrated by the success of the weight-based oracles.
However, the results of the note-based approach imply that the learning/storage
method of the learned knowledge of the oracle is very important. Our implementation
of the weight-only approach showed successful compression of the structured worlds,
while the note-sheet approach was not able to take advantage of the latent structure.
The weight-only oracle is therefore the stronger evidence that
these synthetic worlds contain exploitable latent structure. This mirrors the
distinction between memorizing arbitrary data and learning its generating
process \cite{morris2025memorize}. The note-sheet
oracle has a different strength: it stores object-specific information in a much
smaller observed byte range and, with Qwen2.5-14B, reaches high raw accuracy at
16 KiB. What it does not show in these runs is a detectable advantage on
structured worlds over independent worlds.

\subsection{Medium and Learning Procedure Are Confounded}

One thing to point out is that the fact that the weight-only approach yielded a
stark difference between the world families verifies that the generated worlds
did have exploitable structure. Hence, the failure of the note-sheet approach is
indeed a failure to compress existing structure. This marks a difference in capability
not just in the final oracle format, but in the effectiveness of the learning process.
In our case, the takeaway may not be simply that adapted parameters are a more effective adaptation
medium, but it could also be that gradient updates are a better learning method
than a small number of iterative text updates to an external file. The present experiment therefore identifies an oracle-system
asymmetry, not a clean causal effect of storage medium alone: the adapter system
changes both the medium and the learning algorithm relative to the note-sheet
system.

\section{Limitations}

\subsection{Synthetic Task and Measured Complexity}

Our objects of study are synthetic and very neatly organized, and the complete set of facts about each world is known before training. Moreover, our question templates only cover some of the ways to represent facts about the object, potentially limiting our capacity to test for comprehensive knowledge.

Additionally, the measured storage requirements should not be interpreted as intrinsic lower bounds on the complexity of the worlds. If one oracle requires a given amount of storage to reach a particular accuracy, another specification format or learning procedure might achieve the same accuracy with substantially less. Our results compare the compression achieved by the tested systems; they do not establish an optimal specification size.

\subsection{Model and Harness Generality}

Our experiments use two sizes of Qwen2.5 and one implementation of each oracle type. It is unclear whether the results generalize to other model families, note-writing strategies, adaptation methods, or training configurations. In particular, the plateau in note size and accuracy could indicate that Qwen2.5-7B was unable to make effective use of larger notes, rather than an inherent limitation of storing knowledge in text.

The comparison also does not isolate storage medium from learning procedure. The weight-based oracle learns through gradient updates, whereas the note-based oracle learns through a small number of prompted revisions to an external file. The observed advantage may therefore reflect differences in how the knowledge is learned as well as differences in where it is ultimately stored.

\subsection{Statistical Power}

Statistically speaking, the experiment could benefit from many more samples and
iterations. The \(\DeltaAUC\) values are generally noisy, so the small
number of seeds used for each of the sweeps limits how much inference
we can draw from the data. In the case of our most expensive iterations, the 14B
weight-only runs, the number of test questions additionally makes the accuracy
measurements more noisy, which is reflected in the wider confidence interval.

\appendix

\section{Supplementary Tables and Controls}

\subsection{A1. Headline Results}
\begin{center}
\begin{table}[ht]
\small
\begin{tabularx}{\linewidth}{@{}lllrrlY@{}}
\toprule
Decoder & Oracle & Condition & Seeds & Test \(n\) & \(\DeltaAUC\) & Read \\
\midrule
Qwen2.5-7B & Adapter & Semantic & 3 & 139 & \(+18.5\) pp \([+15.2,+21.8]\) & Large; all seeds positive. \\
Qwen2.5-7B & Note & Semantic & 8 & 100 & \(+1.1\) pp \([-4.3,+6.4]\) & Centered near zero. \\
Qwen2.5-7B & Note & Randomized labels & 3 & 100 & \(+2.4\) pp \([-9.0,+13.8]\) & Noisy. \\
Qwen2.5-14B & Adapter & Semantic & 3 & 32 & \(+16.6\) pp \([-1.8,+35.0]\) & Positive but lower-power. \\
Qwen2.5-14B & Note & Semantic & 8 & 100 & \(+1.5\) pp \([-1.1,+4.1]\) & Near zero. \\
\bottomrule
\end{tabularx}
\caption{Headline structured-minus-independent AUC contrasts.}
\end{table}
\end{center}

\subsection{A2. OpenAI Note Pilot}

Before running the Qwen sweeps, we observed in an OpenAI gpt-5.5
pilot that a stronger note-writing base model could use a few KiB of note budget
to answer independent-world questions accurately. This pilot motivates the
note-sheet idea, but it is not part of the controlled comparison because we did
not run a matched weight-only condition for the same base model.

\begin{center}
\small
\begin{tabular}{@{}rrrrr@{}}
\toprule
Budget (bytes) & Final note bytes & Test \(n\) & Test accuracy & Source \\
\midrule
512 & 461 & 32 & 3.1\% & \path{openai-sweep} \\
1024 & 1019 & 32 & 28.1\% & \path{openai-sweep} \\
2048 & 2041 & 32 & 56.3\% & \path{openai-sweep} \\
4096 & 4024 & 32 & 96.9\% & \path{openai-sweep} \\
2304 & 2304 & 64 & 51.6\% & \path{openai-sweep-2k-4k} \\
2560 & 2560 & 64 & 53.1\% & \path{openai-sweep-2k-4k} \\
3072 & 3072 & 64 & 78.1\% & \path{openai-sweep-2k-4k} \\
3584 & 2638 & 64 & 79.7\% & \path{openai-sweep-2k-4k} \\
\bottomrule
\end{tabular}
\manualtablecaption{Archived OpenAI note pilot on the independent family only.
The two blocks are separate archived sweeps, run at 32 and 64 test questions
respectively.}
\end{center}
The 4 KiB note is the clearest qualitative example. Rather than storing only
full natural-language facts, it begins by defining compact vocabularies for
regions, room types, attributes, owners, and item schemas, then stores one row
per room and a gate-key table. This is still an independent-world run, so it
should be read as evidence that a capable note writer can create a useful
codebook-style external state, not as evidence for cross-family structural
compression. The excerpt below is from
\path{oracle-notes/runs/archived_tmp/openai-sweep/budget-4096/independent/notes/4096.txt}.

\begin{lstlisting}[style=noteexcerpt,caption={Excerpt from the 4 KiB OpenAI pilot note.},label={lst:openai-note-excerpt}]
Reg: BG=Basalt Garden/orange, GC=Glass Coast/blue, IW=Iron Warrens/crimson, AM=Amber Marsh/amber. Room="<adj> <RegName> <Type> <n>". Item="<itemAdj> <regColor> <itemType> <id>", itemType by Type: Sh relic,Ar scroll,Fo gear,Ob lens,Ma token,Ga seed,To charm,Cr urn.
Haz: bm black mold,ab ash bloom,wl white leeches,ef echo fever,sm spore mist,sf silent frost,st static thorns,rg rust gas,sg salt fog.
Room row: n adj Type haz/mat/light/sym/own/itemAdjID.
BG: 1 South Sh bm/vs/pm/sh/FG/North000; 2 Broken Ar ab/vs/gg/og/MC/Broken001; 3 Upper Fo wl/ms/ds/sh/CC/Upper002; ...
Gates: gate n goes room n -> room n+1. Gate names: BG Orange Gate 1-n; GC Blue Gate 2-n; IW Crimson Gate 3-n; AM Amber Gate 4-n.
Fact-id formulas: region name facts BG00000 GC00149 IW00298 AM00447. Room bases BG=1,GC=150,IW=299,AM=448; room n fact=base+9(n-1)+off where off 0 region,1 type,2 hazard,3 material,4 lighting,5 symbol,6 owner,7 contains,8 item located_in.
\end{lstlisting}

\subsection{A3. Adapter Storage Convention}
\begin{center}
\small
\begin{tabular}{@{}llrrr@{}}
\toprule
Model & Layers & Trainable params & 2-byte storage (KiB) & 4-byte storage (KiB) \\
\midrule
Qwen2.5-7B & L1 & 90,112 & 176 & 352 \\
Qwen2.5-7B & L2 & 180,224 & 352 & 704 \\
Qwen2.5-7B & L4 & 360,448 & 704 & 1,408 \\
Qwen2.5-7B & L8 & 720,896 & 1,408 & 2,816 \\
Qwen2.5-7B & L16 & 1,441,792 & 2,816 & 5,632 \\
Qwen2.5-7B & L28 & 2,523,136 & 4,928 & 9,856 \\
\midrule
Qwen2.5-14B & L1 & 89,600 & 175 & 350 \\
Qwen2.5-14B & L2 & 179,200 & 350 & 700 \\
Qwen2.5-14B & L4 & 358,400 & 700 & 1,400 \\
Qwen2.5-14B & L8 & 716,800 & 1,400 & 2,800 \\
Qwen2.5-14B & L16 & 1,433,600 & 2,800 & 5,600 \\
Qwen2.5-14B & L28 & 2,508,800 & 4,900 & 9,800 \\
Qwen2.5-14B & L48 & 4,300,800 & 8,400 & 16,800 \\
\bottomrule
\end{tabular}
\manualtablecaption{Final adapter storage accounting. KiB uses 1024
bytes. The 2-byte column corresponds to fp16/bfloat16 stored LoRA values; the
4-byte column is an fp32 sensitivity check. The shared frozen base model,
optimizer state, training data, and file-format overhead are excluded.}
\end{center}

\subsection{A4. Randomized-Label Note Control}

A three-seed randomized-label note-sheet control run yielded similar results to the semantic names note-sheet runs. Because there is no matched randomized-label
adapter sweep and the semantic note result is already near zero, this control is
diagnostic rather than central to the paper's argument. It is at least a good sanity check that using semantic names does not hurt the oracle's capability.

\end{document}